\documentclass{article}
\usepackage{spconf,amsmath,graphicx,amssymb,bm,xfrac}
\usepackage{multirow}
\usepackage{mathtools}
\usepackage{bbm}

\title{Multiple output samples per input in a single-output Gaussian process}
\name{Jeremy H. M. Wong, Huayun Zhang, and Nancy F. Chen\thanks{This work was supported by the A\textsuperscript{$\star$}STAR Computational Resource Centre through the use of its high performance computing facilities.}}
\address{Institute for Infocomm Research (I\textsuperscript{2}R), A\textsuperscript{$\star$}STAR, Singapore}
\begin{document}
\ninept
\maketitle
\begin{abstract}
The standard Gaussian Process (GP) only considers a single output sample per input in the training set. Datasets for subjective tasks, such as spoken language assessment, may be annotated with output labels from multiple human raters per input. This paper proposes to generalise the GP to allow for these multiple output samples in the training set, and thus make use of available output uncertainty information. This differs from a multi-output GP, as all output samples are from the same task here. The output density function is formulated to be the joint likelihood of observing all output samples, and latent variables are not repeated to reduce computation cost. The test set predictions are inferred similarly to a standard GP, with a difference being in the optimised hyper-parameters. This is evaluated on speechocean762, showing that it allows the GP to compute a test set output distribution that is more similar to the collection of reference outputs from the multiple human raters.
\end{abstract}
\begin{keywords}
Gaussian process, subjective, uncertainty, inter-rater agreement, spoken language assessment
\end{keywords}
\section{Introduction}

The Gaussian Process (GP) \cite{rasmussen2006} expresses a distributional uncertainty \cite{malinin2018} that naturally increases for inputs further away from the training data. It may also be interpreted as marginalising over functions \cite{rasmussen2006}, which reduces the influence of model uncertainty \cite{kendall2017,mackay1992a}. However, a GP does not consider whether outputs are similar for proximate inputs, and thus may lack in modelling data uncertainty \cite{kendall2017,xu2021}. Better uncertainty estimation may be especially desirable for subjective tasks, such as Spoken Language Assessment (SLA), where multiple human annotators may provide differing output labels for the same input \cite{junbozhang2021}. A collection of human annotations for the same input may be interpreted as a reference of data uncertainty that an automatic model should also aim to compute. In such settings, the uncertainties expressed by the model and the human annotators can be explicitly compared.

The standard GP assumes that each input in the training set is paired with only one output, treated as the ground truth. This paper proposes to extend the GP formulation, to use available information of data uncertainty from multiple output samples for each input in the training set. A naive approach is to repeat the inputs for each output, but this greatly increases the computational cost. A proposed omission of repeated latent variables avoids this expense. All outputs here are from the same task, while the multi-output GP \cite{yu2005,bonilla2007} instead considers a multi-task framework \cite{caruana1997}.

The aim of SLA is to assign a score related to oral proficiency, to a speech input. Examples of proficiency aspects that are often assessed include pronunciation accuracy, intonation, fluency, prosody, task completion, and topic relevance. As previously mentioned, SLA is a subjective task, where differing scores are often assigned by multiple expert raters for the same input. The diversity of scores can be reduced by increasing the coverage of rules in a rubric. However, the choice of rules may not agree with the needs of diverse users. This paper instead embraces subjectivity, by using the diversity of reference scores to better predict output uncertainty. This uncertainty information may allow the system to better calibrate its feedback to the user, by not penalising a student for a low score that raters would have disagreed upon, and instead seek clarification or human intervention. This also reduces the need for a strict rubric, thereby allowing better generalisation to diverse user needs.

\section{Gaussian process regression}

When given a training set of $N$ input feature vectors of dimension $D$, $\mathbf{X}\in\mathbb{R}^{N\times D}$, a GP places a jointly Gaussian prior over latent variables, $\bm{f}\in\mathbb{R}^N$, as $p(\bm{f}|\mathbf{X})=\mathcal{N}(\bm{f};\bm{0},\mathbf{K}^{\mathbf{X}\mathbf{X}})$. The covariance of the latent variables is defined as a pair-wise similarity between the inputs, computed by the kernel, $\mathbf{K}$. Here, the squared exponential kernel is used, with elements defined as $k_{ij}^{\mathbf{X}\mathbf{X}^\prime}=s^2\exp[-\frac{1}{2l^2}(\bm{x}_i-\bm{x}_j^\prime)^\top(\bm{x}_i-\bm{x}_j^\prime)]$, where $i$ and $j$ are the data point indexes, $l$ is a length hyper-parameter, and $s$ is a scale hyper-parameter. The GP makes the assumption that the outputs, $\bm{y}\in\mathbb{R}^N$, are conditionally independent of the inputs when given the latent variables. The outputs are Gaussian distributed, $p(\bm{y}|\bm{f})=\mathcal{N}(\bm{y};\bm{f},\sigma^2\mathbf{I})$, about a mean of the latent variable, with an observation noise hyper-parameter, $\sigma$, where $\mathbf{I}$ is the identity matrix.

Training a GP involves estimating the hyper-parameters $s$, $l$, and $\sigma$. One approach is to maximise the marginal log-likelihood of the training data, $\mathcal{F}=\log p(\bm{y}^\text{ref}|\mathbf{X})$, where $\bm{y}^\text{ref}$ are the reference outputs. The marginal likelihood is
\begin{equation}
p\left(\bm{y}\middle|\mathbf{X}\right)=\int p\left(\bm{y}\middle|\bm{f}\right)p\left(\bm{f}\middle|\mathbf{X}\right)d\bm{f}=\mathcal{N}\left(\bm{y};\bm{0},\mathbf{K}^{\mathbf{X}\mathbf{X}}\!+\!\sigma^2\mathbf{I}\right)\!\!.
\end{equation}
Gradient-based optimisation of the hyper-parameters requires inverting $\mathbf{K}^{\mathbf{X}\mathbf{X}}+\sigma^2\mathbf{I}$ \cite{rasmussen2006}, with a number of operations scaling as $\mathcal{O}(N^3)$. Algorithms such as \cite{petkovic2009} reduce the polynomial power, but still require more than $\mathcal{O}(N^2)$ operations. For evaluation, outputs, $\widehat{\bm{y}}$, are predicted from test inputs, $\widehat{\mathbf{X}}$, by first computing a posterior over predicted latent variables,
\begin{equation}
p\left(\widehat{f}\middle|\widehat{\bm{x}},\bm{y},\mathbf{X}\right)=\frac{p\left(\widehat{f},\bm{y}\middle|\widehat{\bm{x}},\mathbf{X}\right)}{p\left(\bm{y}\middle|\mathbf{X}\right)}\label{eq:predictive_bayes_rule}=\mathcal{N}\left(\widehat{f};\widehat{\mu},\widehat{v}\right),
\end{equation}
where $\widehat{\mu}={\bm{k}^{\mathbf{X}\widehat{\bm{x}}}}^\top[\mathbf{K}^{\mathbf{X}\mathbf{X}}+\sigma^2\mathbf{I}]^{-1}\bm{y}$ and $\widehat{v}=k^{\widehat{\bm{x}}\widehat{\bm{x}}}-{\bm{k}^{\mathbf{X}\widehat{\bm{x}}}}^\top[\mathbf{K}^{\mathbf{X}\mathbf{X}}+\sigma^2\mathbf{I}]^{-1}\bm{k}^{\mathbf{X}\widehat{\bm{x}}}$, which similarly to training, also requires computing $[\mathbf{K}^{\mathbf{X}\mathbf{X}}+\sigma^2\mathbf{I}]^{-1}$. The posterior for the test output is then $p(\widehat{y}|\widehat{\bm{x}},\bm{y},\mathbf{X})=\mathcal{N}(\widehat{y};\widehat{\mu},\widehat{v}+\sigma^2)$. Finally, the predicted scalar output for each test data point, $i$, can be inferred from this, using a decision rule of, for example, choosing the mean, $\widehat{y}_i^\star=\widehat{\mu}_i$. The mean, mode, and median are equivalent because of the Gaussian symmetry and uni-modality.

\section{Multiple output samples}
\label{sec:multiple_output_samples}

In the standard GP, each input is paired with a single output sample. In SLA, using a GP with only a scalar output label as the reference \cite{vandalen2015} may forego information about the inter-rater uncertainty. This paper proposes extending the GP to use the multiple output samples in the training set. Let each training input be associated with $R$ output reference samples, $\bm{y}_i^\text{ref}=[y_{i1}^\text{ref},\cdots,y_{iR}^\text{ref}]$, where $y_{ir}^\text{ref}$ is the $r$th reference output sample for the $i$th data point. It is assumed for simplicity that the number of output samples is the same for all inputs, but the formulation can be extended for varying numbers of raters \cite{wong2023asru}. The outputs are stacked across all data points to yield a matrix $\mathbf{Y}^\text{ref}\in\mathbb{R}^{N\times R}$.

A naive approach to accommodate multiple output samples is to flatten $\mathbf{Y}^\text{ref}$ into a vector and repeat the inputs for each output sample. This will yield a $\mathbb{R}^{NR\times NR}$ kernel of $\widetilde{\mathbf{K}}=\mathbf{K}(\mathbf{X},\mathbf{X})\otimes\mathbf{1}^{(R)}$, where $\mathbf{1}^{(R)}$ is a $\mathbb{R}^{R\times R}$ matrix of 1s and $\otimes$ represents a Kronecker (tensor) product. There is no need to explicitly invert this non-full-rank kernel, as only the inverse of the full-rank $\widetilde{\mathbf{K}}+\sigma^2\mathbf{I}$ is needed. Inversion requires a number of operations that scales as $\mathcal{O}(N^3R^3)$, as opposed to $\mathcal{O}(N^3)$ when using only a single output sample. Computational constraints may then impose the need for sparse approximation \cite{quinonero-candela2005}, even for modest training set sizes.


When using this repeated kernel, the latent variables prior is $p(\widetilde{\bm{f}}|\mathbf{X})=\mathcal{N}(\widetilde{\bm{f}}|\bm{0},\widetilde{\mathbf{K}})$. The repeated structure of $\widetilde{\mathbf{K}}$ causes all latent variables associated with the same training data point to be perfectly correlated, and thus have the same value. It seems computationally redundant to separately model the repeated latent variables. This paper proposes that there is no need to separately compute each of the repetitions, and to instead only express a single instance of each latent variable per training data point. The joint density of the multiple output samples is then conditioned on these non-repeated latent variables, $p(\mathbf{Y}|\bm{f})$. The diagonal covariance in the output density function of a standard GP implies that the outputs from different training data points are conditionally independent of each other when given the latent variables. This conditional independence, applied to the case of repeated inputs in the naive approach, yields in this proposed approach a joint output density that can be factorised as
\begin{align}
p\left(\mathbf{Y}\middle|\bm{f}\right)=&\prod_{r=1}^R\mathcal{N}\left(\bm{y}_r;\bm{f},\sigma^2\mathbf{I}\right)\label{eq:output_density_mean_std}\\
=&\,\mathcal{N}\left(\overline{\bm{\mu}};\bm{f},\frac{\sigma^2}{R}\mathbf{I}\right)\mathcal{N}\left(\overline{\bm{\eta}};\bm{0},\frac{\sigma^2}{R}\mathbf{I}\right)\frac{\left(2\pi\sigma^2\right)^{\left(2-R\right)\frac{N}{2}}}{R^N},\notag
\end{align}
where $\overline{\bm{\mu}}=\frac{1}{R}\sum_r\bm{y}_r$ and $\overline{\eta}_i=\sqrt{\frac{1}{R}\sum_r(y_{ir}-\overline{\mu}_i)^2}$, and $\bm{y}_r\in\mathbb{R}^N$ is a vector of the $r$th output sample for all inputs. The factorisation and diagonal covariance in \eqref{eq:output_density_mean_std} assume that the multiple output samples for the same input and across different inputs are independent of each other, when given the latent variable. As a result, the output density does not depend on the order of the multiple output samples for each input. This can be explicitly seen in the re-parameterisation of the multiple outputs as the empirical mean $\overline{\bm{\mu}}$ and empirical biased standard deviation $\overline{\bm{\eta}}$, both of which are independent of the ordering of the output samples. Also, \eqref{eq:output_density_mean_std} may seem unnormalised over $\overline{\bm{\mu}}$ and $\overline{\bm{\eta}}$. However, it should be noted that $p(\mathbf{Y}|\bm{f})$ is expected to sum to one when integrated over $\mathbf{Y}$, and not over $\overline{\bm{\mu}}$ and $\overline{\bm{\eta}}$.


The marginal likelihood can be computed from the non-redundant prior and the output density of \eqref{eq:output_density_mean_std}, as
\begin{equation}
p\!\left(\mathbf{Y}\middle|\mathbf{X}\right)\!=\!\!\!\int \!\!p\!\left(\mathbf{Y}\middle|\bm{f}\right)p\!\left(\bm{f}\middle|\mathbf{X}\right)d\bm{f}\!=\!\mathcal{N}\!\left(\!\overline{\bm{\mu}};\!\bm{0},\!\mathbf{K}^{xx}\!\!+\!\!\frac{\sigma^2}{R}\mathbf{I}\!\right)g\!\left(\mathbf{Y}\right)\!\!,
\end{equation}
where $g(\mathbf{Y})=\mathcal{N}(\overline{\bm{\eta}};\bm{0},\frac{\sigma^2}{R}\mathbf{I})(2\pi\sigma^2)^{(2-R)\frac{N}{2}}R^{-N}$. Analogously to the standard GP, and equivalently to the naive approach, the hyper-parameters here can be optimised by maximising the joint marginal log-likelihood of all output samples, $\mathcal{F}_\text{joint}=\log p(\mathbf{Y}^\text{ref}|\mathbf{X})$. Unlike a naive implementation, the computational cost of gradient-based optimisation of the proposed method scales as $\mathcal{O}(N^3)$, similarly to a standard GP with a single output sample.

When performing evaluation, the predicted output can be inferred from the posterior. First, the joint density between the training set outputs and the test latent variables needs to be computed. When the training set comprises multiple output samples per input, this paper again proposes that there is no need to separately model redundant latent variables, by expressing this joint density as
\begin{align}
p\left(\widehat{f},\mathbf{Y}\middle|\widehat{\bm{x}},\mathbf{X}\right)&=\int p\left(\widehat{f},\bm{f}\middle|\widehat{\bm{x}},\mathbf{X}\right)p\left(\mathbf{Y}\middle|\bm{f}\right)d\bm{f}\\
&=\mathcal{N}\left(\left[\begin{matrix}\widehat{f}\\\overline{\bm{\mu}}\end{matrix}\right];\bm{0},\left[\begin{matrix}k^{\widehat{\bm{x}}\widehat{\bm{x}}}&{\bm{k}^{\mathbf{X}\widehat{\bm{x}}}}^\top\\\bm{k}^{\mathbf{X}\widehat{\bm{x}}}&\mathbf{K}^{\mathbf{X}\mathbf{X}}+\frac{\sigma^2}{R}\mathbf{I}\end{matrix}\right]\right)g\left(\mathbf{Y}\right)\!.\notag
\end{align}
The test set latent variables posterior is
\begin{equation}
p\left(\widehat{f}\middle|\widehat{\bm{x}},\mathbf{Y},\mathbf{X}\right)=\frac{p\left(\widehat{f},\mathbf{Y}\middle|\widehat{\bm{x}},\mathbf{X}\right)}{p\left(\mathbf{Y}\middle|\mathbf{X}\right)}=\mathcal{N}\left(\widehat{f};\breve{\mu},\breve{v}\right),
\end{equation}
analogously to \eqref{eq:predictive_bayes_rule}, where here, $\breve{\mu}={\bm{k}^{\mathbf{X}\widehat{\bm{x}}}}^\top[\mathbf{K}^{\mathbf{X}\mathbf{X}}+\frac{\sigma^2}{R}\mathbf{I}]^{-1}\overline{\bm{\mu}}$ and $\breve{v}=k^{\widehat{\bm{x}}\widehat{\bm{x}}}-{\bm{k}^{\mathbf{X}\widehat{\bm{x}}}}^\top[\mathbf{K}^{\mathbf{X}\mathbf{X}}+\frac{\sigma^2}{R}\mathbf{I}]^{-1}\bm{k}^{\mathbf{X}\widehat{\bm{x}}}$. The test output posterior is then $p(\widehat{y}|\widehat{\bm{x}},\mathbf{Y},\mathbf{X})=\mathcal{N}(\widehat{y};\breve{\mu},\breve{v}+\sigma^2)$, from which an output can be chosen. The matrix inverted here is of size $\mathbb{R}^{N\times N}$, as opposed to $\mathbb{R}^{NR\times NR}$ in the naive repetition approach. The number of operations thus scales as $\mathcal{O}(N^3)$, similarly to a standard GP with a single output sample.

The covariance of the standard and proposed posteriors are independent of the training set outputs, and only depend on the training set inputs and the hyper-parameters. As such, information about the proximity between the multiple output samples for the same input in the training set does not influence the resulting posterior covariance. Furthermore, whether the training set output values are near or far from each other, for two inputs that are near as measured by the kernel, does not influence the resulting posterior covariance for test inputs near to these training inputs. The posterior covariance is only affected by the distance between the test and training set inputs. This implies that a GP may be more suited for computing distributional uncertainty, and less so for the data uncertainty \cite{xu2021} that may be more relevant to the uncertainty expressed between the raters. Work in \cite{wong2023asru} extends this framework to allow the GP to take training set outputs into account for the posterior covariance.

The predictive mean, $\breve{\mu}$, is dependent on the mean of the multiple output samples, $\overline{\bm{\mu}}$, and independent of $\overline{\bm{\eta}}$. This is similar to a standard GP, when using a mean combination of the multiple rater scores to get the reference. As mentioned previously, the predictive covariance is independent of the training set outputs. Thus, during inference, the remaining difference between the standard GP and the proposal is in the hyper-parameters. The proposal optimises the hyper-parameters by maximising the joint marginal log-likelihood of all training set outputs, while the standard approach maximises the marginal log-likelihood of the combined output reference.

\section{Experiments}


Experiments were performed on speechocean762 \cite{junbozhang2021}, comprising training and test sets, each with 2500 sentences and 125 disjoint speakers. The native Mandarin speakers read the sentences in English. Each sentence is annotated by 5 raters with a variety of score types at different linguistic levels, but only the sentence-level pronunciation accuracy was used, with an integer score ranging between 0 and 10. The scalar reference scores were computed as a mean of the multiple rater scores, differing from the median used in \cite{junbozhang2021}. The mean may yield a baseline that better matches \eqref{eq:output_density_mean_std}. Following \cite{junbozhang2021}, the predicted and combined reference scores were rounded to integers before computing the Pearson's Correlation Coefficient (PCC) and Mean Squared Error (MSE). These metrics only compare scalar references and hypotheses. A model with a probabilistic output can also be assessed on how well it matches the distribution of scores from multiple raters, and thus its ability to predict the inter-rater uncertainty. The reference distribution over integer scores, $c$, can be expressed as a mixture of Kronecker $\delta$-functions at each rater's score, $P^\text{ref}(c|\widehat{\bm{x}}_i)=\frac{1}{R}\sum_r\delta(c,y^\text{ref}_{ir})$, showing the fraction of raters who assigned each score. This Monte Carlo approximation to a Bayesian NN \cite{mackay1992a} marginalises over the human equivalent of model uncertainty, yielding a purer reference of data uncertainty. The model's output posterior can be discretised as $P(c|\widehat{\bm{x}}_i,\mathbf{Y},\mathbf{X})=\frac{\int_{c-0.5}^{c+0.5} p(\widehat{y}_i|\widehat{\bm{x}}_i,\mathbf{Y},\mathbf{X})d\widehat{y}_i}{\sum_{c^\prime}\int_{c^\prime-0.5}^{c^\prime+0.5} p(\widehat{y}_i^\prime|\widehat{\bm{x}}_i,\mathbf{Y},\mathbf{X})d\widehat{y}_i^\prime}$, to express the probability of each integer score. The reference and hypothesised distributions can then be compared using a discrete Kullback-Leibler (KL) divergence between $P^\text{ref}(c|\widehat{\bm{x}}_i)$ and $P(c|\widehat{\bm{x}}_i,\mathbf{Y},\mathbf{X})$. Unlike a continuous KL divergence, this is lower-bounded by 0, easing interpretability, and follows \cite{junbozhang2021} by rounding scores before evaluating.

Statistical significance, $\rho$, for MSE and KL divergence was computed using a two-tailed paired $t$-test. The PCC is not easily expressed as a sum over data points, hindering application of the central limit theorem. Instead, the $Z_1^\star$ approach was used \cite{steiger1980}. This first computes an approximately normally distributed transformation \cite{dunn1969} from the two PCCs being compared, then the two-tailed cumulative density of this transformed variable yields the significance.

The feature extraction followed \cite{huayunzhang2021}. First, a hybrid speech recognition model force-aligned the audio with the transcriptions. This was trained using Kaldi \cite{kaldi} on 960 hours Librispeech \cite{panayotov2015} using cross-entropy, following \cite{junbozhang2021}. From the forced alignment, goodness of pronunciation \cite{witt2000}, log phone posterior \cite{hu2015}, log posterior ratio \cite{hu2015}, tempo \cite{huayunzhang2021}, and pitch \cite{ghahremani2014} features were extracted. A continuous skip-gram model \cite{mikolov2013}, with a 32-node recurrent NN \cite{elman1990} hidden layer, was trained on the training set non-silence phone sequences, and used to extract phone embeddings \cite{huayunzhang2021}. These were concatenated to form one feature vector per phone.

It may not be trivial to allow a GP to use sequential inputs. Work in \cite{lodhi2000} designs kernels to operate on sequences, while \cite{vandalen2015} extracts hand-crafted sentence-level features for SLA. This paper computed sentence-level inputs using a NN feature extractor. The sequence of features, with length equal to the number of phones in the sentence, was fed into a Bidirectional Long Short-Term Memory (BLSTM) layer \cite{graves2005}, with 32 nodes per direction. The output was then pooled across all phones in the sequence with equal weights, and fed through linear and sigmoid layers. The output was then scaled to the bounds of the output score range. The NN was trained by minimising the MSE toward the combined reference score. The activations after the pooling layer formed inputs for the GP \cite{salakhutdinov2007}. The GP and NN feature extractor were not jointly fine-tuned, as doing so may overfit easily \cite{ober2021}. Principle component analysis whitening, estimated on the training set, was applied to these features to better abide by the diagonal covariance assumption of the scalar length hyper-parameter in the kernel. Dropout \cite{srivastava2014} with an omission probability of 60\% was used before the BLSTM and linear layers. For NN training, a 10\% validation set was held out without speaker disjointment.

\begin{table}[t]
\centering
\caption{Using training set scores from multiple raters in a GP}
\label{tab:multioutput}
\begin{tabular}{l|cc|c|c}
\hline
Model&PCC$\uparrow$&MSE$\downarrow$&KL$\downarrow$&Inference time (s)$\downarrow$\\
\hline\hline
GP\textsubscript{base}&0.710&1.149&3.10&136 $\pm$ 3\\
\hline
GP\textsubscript{repeat}&\multirow{2}{*}{0.713}&\multirow{2}{*}{1.136}&\multirow{2}{*}{0.85}&8476 $\pm$ 510\\
GP\textsubscript{joint}&&&&138 $\pm$ 4\\
\hline
\end{tabular}
\end{table}

The experiment assesses the proposed extension of allowing the GP to use the separate output scores from multiple raters in the training set, referred to as GP\textsubscript{joint}. This is compared against using only a single output reference in GP\textsubscript{base}, and also against a naive multiple output samples extension of repeating the training set input features in GP\textsubscript{repeat}. The GP\textsubscript{joint} and GP\textsubscript{repeat} hyper-parameters were optimised by maximising $\mathcal{F}_\text{joint}$. In table \ref{tab:multioutput}, the PCC and MSE results may suggest that the proposed GP\textsubscript{joint} approach yields performance gains over GP\textsubscript{base}, but these may not be significant, with $\rho_\text{PCC}=0.065$ and $\rho_\text{MSE}=0.036$. The PCC and MSE compare the prediction against the mean combined reference, assuming that this scalar reference is correct, and do not consider how well the model's output uncertainty matches the uncertainty between the raters. The KL divergence is used to assess this uncertainty matching. The results show that using the separate scores from multiple training set raters improves the KL divergence of the GP, with $\rho_\text{KL}<0.001$. This suggests that the proposed extension allows the GP's output distribution to better match that represented by the collection of scores from the multiple raters. Such a more accurate uncertainty prediction may better inform the type of feedback that should be given to the user.

The computational saving of GP\textsubscript{joint} over GP\textsubscript{repeat} was assessed by repeating inference 10 times on a 48 core Intel Xeon Platinum 8268 2.90GHz CPU using the Numpy pseudo-inverse implementation. The mean and standard deviation of the total time to infer the whole test set accumulated over all threads are shown in the right-side column in table \ref{tab:multioutput}. Inference using a GP can be run efficiently by pre-computing the matrix inverse before any test data points are seen. It therefore seems more interpretable to view the total time across the whole test set, rather than the time per data point or real-time factor, assuming that the computation time is dominated by matrix inversion. The results show that GP\textsubscript{joint} is indeed faster to infer from than GP\textsubscript{repeat}. GP\textsubscript{joint} and GP\textsubscript{base} require comparable durations to infer from. Although not shown here, this computational saving also benefits the matrix inversion computation during training.

\section{Conclusion}

This work extends the GP to allow for multiple output samples per input in the training set with computational efficiency. The hyper-parameters can be optimised by maximising the joint marginal likelihood of all output samples. Inference is performed in a fairly similar way to a standard GP, with the main deviation being in the hyper-parameters that are optimised differently. This allows the posterior to better match the uncertainty expressed by the multiple raters. Such information may better inform the type of feedback given to the user.

\bibliographystyle{IEEEbib}
\bibliography{strings,refs}

\begin{thebibliography}{10}

\bibitem{rasmussen2006}
C.~E. Rasmussen and C.~K.~I. Williams,
\newblock {\em Gaussian processes for machine learning},
\newblock MIT Press, 2006.

\bibitem{malinin2018}
A.~Malinin and M.~Gales,
\newblock ``Predictive uncertainty estimation via prior networks,''
\newblock in {\em NeurIPS}, Montr\'{e}al, Canada, Dec 2018, pp. 7047--7058.

\bibitem{kendall2017}
A.~Kendall and Y.~Gal,
\newblock ``What uncertainties do we need in {B}ayesian deep learning for
  computer vision?,''
\newblock in {\em NIPS}, Long Beach, USA, Dec 2017, pp. 5574--5584.

\bibitem{mackay1992a}
D.~J.~C. MacKay,
\newblock ``Bayesian interpolation,''
\newblock {\em Neural Computation}, vol. 4, no. 3, pp. 415--447, May 1992.

\bibitem{xu2021}
B.~Xu, R.~Kuplicki, S.~Sen, and M.~P. Paulus,
\newblock ``The pitfalls of using {G}aussian process regression for normative
  modeling,''
\newblock {\em PLoS ONE}, vol. 16, no. 9, Sep 2021.

\bibitem{junbozhang2021}
J.~Zhang, Z.~Zhang, Y.~Wang, Z.~Yan, Q.~Song, Y.~Huang, K.~Li, D.~Povey, and
  Y.~Wang,
\newblock ``speechocean762: an open-source non-native {E}nglish speech corpus
  for pronunciation assessment,''
\newblock in {\em Interspeech}, Brno, Czechia, Aug 2021, pp. 3710--3714.

\bibitem{yu2005}
K.~Yu, V.~Tresp, and A.~Schwaighofer,
\newblock ``Learning {G}aussian processes from multiple tasks,''
\newblock in {\em ICML}, Bonn, Germany, Aug 2005, pp. 1012--1019.

\bibitem{bonilla2007}
E.~V. Bonilla, K.~M.~A. Chai, and C.~K.~I. Williams,
\newblock ``Multi-task {G}aussian process prediction,''
\newblock in {\em NIPS}, Vancouver, Canada, Dec 2007, pp. 153--160.

\bibitem{caruana1997}
R.~Caruana,
\newblock ``Multitask learning,''
\newblock {\em Machine Learning}, vol. 28, no. 1, pp. 41--75, Jul 1997.

\bibitem{petkovic2009}
M.~D. Petkovi\'{c} and P.~S. Stanimirovi\'{c},
\newblock ``Generalized matrix inversion is not harder than matrix
  multiplication,''
\newblock {\em Journal of Computational and Applied Mathematics}, vol. 230, no.
  1, pp. 270--282, Aug 2009.

\bibitem{vandalen2015}
R.~C. van Dalen, K.~M. Knill, and M.~J.~F. Gales,
\newblock ``Automatically grading learners' {E}nglish using a {G}aussian
  process,''
\newblock in {\em SLaTE}, Leipzig, Germany, Sep 2015, pp. 7--12.

\bibitem{wong2023asru}
J.~H.~M. Wong, H.~Zhang, and N.~F. Chen,
\newblock ``Variational {G}aussian process data uncertainty,''
\newblock in {\em ASRU}, Taipei, Taiwan, Dec 2023.

\bibitem{quinonero-candela2005}
J.~Qui$\tilde{\text{n}}$onero-Candela and C.~E. Rasmussen,
\newblock ``A unifying view of sparse approximate {G}aussian process
  regression,''
\newblock {\em JMLR}, vol. 6, no. 65, pp. 1939--1959, Dec 2005.

\bibitem{steiger1980}
J.~H. Steiger,
\newblock ``Tests for comparing elements of a correlation matrix,''
\newblock {\em Psychological Bulletin}, vol. 87, no. 2, pp. 245--251, 1980.

\bibitem{dunn1969}
O.~J. Dunn and V.~Clark,
\newblock ``Correlation coefficients measured on the same individuals,''
\newblock {\em Journal of the American Statistical Association}, vol. 64, no.
  325, pp. 366--377, Mar 1969.

\bibitem{huayunzhang2021}
H.~Zhang, K.~Shi, and N.~F. Chen,
\newblock ``Multilingual speech evaluation: case studies on {English}, {Malay}
  and {Tamil},''
\newblock in {\em Interspeech}, Brno, Czechia, Aug 2021, pp. 4443--4447.

\bibitem{kaldi}
D.~Povey, A.~Ghoshal, G.~Boulianne, L.~Burget, O.~Glembek, N.~Goel,
  M.~Hannemann, P.~Motl\'{i}\v{c}ek, Y.~Qian, P.~Schwarz, J.~Silovsk\'{y},
  G.~Stemmer, and K.~Vesel\'{y},
\newblock ``The {Kaldi} speech recognition toolkit,''
\newblock in {\em ASRU}, Hawaii, USA, Dec 2011.

\bibitem{panayotov2015}
V.~Panayotov, G.~Chen, D.~Povey, and S.~Khudanpur,
\newblock ``Librispeech: an {ASR} corpus based on public domain audio books,''
\newblock in {\em ICASSP}, Brisbane, Australia, Apr 2015, pp. 5206--5210.

\bibitem{witt2000}
S.~M. Witt and S.~J. Young,
\newblock ``Phone-level pronunciation scoring and assessment for interactive
  language learning,''
\newblock {\em Speech Communication}, vol. 30, no. 2-3, pp. 95--108, Feb 2000.

\bibitem{hu2015}
W.~Hu, Y.~Qian, F.~K. Soong, and Y.~Wang,
\newblock ``Improved mispronunciation detection with deep neural network
  trained acoustic models and transfer learning based logistic regression
  classifiers,''
\newblock {\em Speech Communication}, vol. 67, pp. 154--166, Mar 2015.

\bibitem{ghahremani2014}
P.~Ghahremani, B.~BabaAli, D.~Povey, K.~Riedhammer, J.~Trmal, and S.~Khudanpur,
\newblock ``A pitch extraction algorithm tuned for automatic speech
  recognition,''
\newblock in {\em ICASSP}, Florence, Italy, May 2014, pp. 2494--2498.

\bibitem{mikolov2013}
T.~Mikolov, K.~Chen, G.~Corrado, and J.~Dean,
\newblock ``Efficient estimation of word representations in vector space,''
\newblock in {\em ICLR}, Scottsdale, USA, May 2013.

\bibitem{elman1990}
J.~L. Elman,
\newblock ``Finding structure in time,''
\newblock {\em Cognitive Science}, vol. 14, no. 2, pp. 179--211, Apr 1990.

\bibitem{lodhi2000}
H.~Lodhi, J.~Shawe-Taylor, N.~Cristianini, and C.~Watkins,
\newblock ``Text classification using string kernels,''
\newblock in {\em NIPS}, Denver, USA, Nov 2000, pp. 563--569.

\bibitem{graves2005}
A.~Graves and J.~Schmidhuber,
\newblock ``Framewise phoneme classification with bidirectional {LSTM}
  networks,''
\newblock in {\em IJCNN}, Montreal, Canada, Jul 2005, pp. 2047--2052.

\bibitem{salakhutdinov2007}
R.~Salakhutdinov and G.~E. Hinton,
\newblock ``Using deep belief nets to learn covariance kernels for {G}aussian
  processes,''
\newblock in {\em NIPS}, Vancouver, Canada, Dec 2007, pp. 1249--1256.

\bibitem{ober2021}
S.~W. Ober, C.~E. Rasmussen, and M.~van~der Wilk,
\newblock ``The promises and pitfalls of deep kernel learning,''
\newblock in {\em UAI}, Jul 2021, pp. 1206--1216.

\bibitem{srivastava2014}
N.~Srivastava, G.~E. Hinton, A.~Krizhevsky, I.~Sutskever, and R.~Salakhutdinov,
\newblock ``Dropout: a simple way to prevent neural networks from
  overfitting,''
\newblock {\em Journal of Machine Learning Research}, vol. 15, no. 56, pp.
  1929--1958, Jun 2014.

\end{thebibliography}

\end{document}